\documentclass{acm_proc_article-sp}
\usepackage{graphicx}
\usepackage{times}
\usepackage{subfigure}
\usepackage{enumitem}
\usepackage{balance}
\usepackage{url}
\usepackage{enumitem}
\setenumerate{itemsep=3pt,topsep=0pt,parsep=0pt,partopsep=0pt,leftmargin=1.5pc}
\usepackage{common}

\begin{document}

\title{Unveiling the Political Agenda of the European Parliament Plenary: A Topical Analysis}

\numberofauthors{2}
\author{
  \alignauthor Derek Greene\\
    \affaddr{School of Computer Science \& Informatics}\\
    \affaddr{University College Dublin, Ireland}\\
    \email{derek.greene@ucd.ie}
  \alignauthor James P. Cross\\
    \affaddr{School of Politics \& International Relations}\\
    \affaddr{University College Dublin, Ireland}\\
    \email{james.cross@ucd.ie}
}

\maketitle

\begin{abstract}
This study analyzes political interactions in the European Parliament (EP) by considering how the political agenda of the plenary sessions has evolved over time and the manner in which Members of the European Parliament (MEPs) have reacted to external and internal stimuli when making Parliamentary speeches. It does so by considering the context in which speeches are made, and the content of those speeches. To detect latent themes in legislative speeches over time, speech content is analyzed using a new dynamic topic modeling method, based on two layers of matrix factorization. This method is applied to a new corpus of all English language legislative speeches in the EP plenary from the period 1999-2014. Our findings suggest that the political agenda of the EP has evolved significantly over time, is impacted upon by the committee structure of the Parliament, and reacts to exogenous events such as EU Treaty referenda and the emergence of the Euro-crisis have a significant impact on what is being discussed in Parliament. 

\end{abstract}

\section{Introduction}
\label{sec:intro}

The plenary sessions of the European Parliament (EP) are one of the most important arenas in which European representatives can air questions, express criticisms and take policy positions to influence EU politics. Indeed the plenary of the Parliament represents the closest that the European Union (EU) gets to engaging in the core democratic process of publicly-aired democratic debate. As a result, understanding how Members of the European Parliament (MEPs) express themselves in plenary, and investigating how the political discussions evolve and respond to internal and external stimuli is a fundamentally important undertaking. 


In recent years, there has been a concurrent explosion of online records detailing the content of MEP speeches, and the development of data mining techniques capable of extracting latent patterns in content across sets of these speeches. This allows us for the first time to investigate the plenary agenda of the Parliament in a holistic and rigorous manner. One approach to tracking the political attention of political figures has been to apply topic modeling algorithms to large corpora of political texts, such as parliamentary speeches of the U.S. Senate \cite{quinn2010analyze}. These algorithms seek to distill the latent thematic patterns in a corpus of speeches \cite{blei03lda}, and can be used to improve the transparency of the political process by providing a macro-level overview of the activities and agendas of politicians in a time- and resource-efficient manner. This type of overview would otherwise be unavailable due to the time and resource costs associated with manually hand-coding such a large-scale corpus.

This paper takes up the challenge of extracting latent thematic patterns in political speeches by developing a suitable dynamic topic modeling method\footnote{\url{https://github.com/derekgreene/dynamic-nmf}} to investigate how the plenary agenda of the EP has changed over three parliamentary terms (1999--2014), based on the analysis of a corpus of 210,247 speeches from 1,735 MEPs across the 28 EU member states. The method described in \refsec{sec:methods} involves applying two layers of Non-negative Matrix Factorization (NMF) topic modeling \cite{lee99nmf}. Firstly, the corpus of speeches is divided into distinct segments or \emph{time windows}, on which low-level \emph{window topics} are identified by applying NMF. Secondly, the topics from each window are represented as a combined matrix of ``topic documents''. By applying NMF to this new representation, we can identify high-level \emph{dynamic topics} which potentially span many time windows. This process allows us to explore parliamentary activity both at a granular level and over multiple parliamentary terms. In addition, we can relate these dynamic topics to the original speakers, allowing us to identify the topics to which individual MEPs are paying most attention.  

Applying our proposed topic modeling methodology reveals the breadth of policy areas covered by the EP, and the results presented later in \refsec{sec:eval} indicate that the political agenda of the Parliament has evolved significantly across the three parliamentary terms considered here. By examining a number of topic case studies, ranging from the Euro-crisis to EU treaty changes, we can identify the relationship between the evolution of these dynamic topics and the exogenous events driving them. By using external data sources, we can also confirm the semantic and construct validity of these topics. In order to explain some of the patterns in speech making we observe, we conclude the paper with an exploration of the determinants of MEP speech-making behavior on the topics detected by our topic model. To provide access to the results of the project to interested parties, we make a browsable version available online\footnote{\url{http://erdos.ucd.ie/europarl}}. This website provides a greater level of transparency into the activities of the EP as a functioning democratic institution.


\section{Related Work}
\label{sec:related}

\subsection{European Parliament}
 
The most prominent forms of MEP behavior that have been examined in the existing literature include the expression of policy positions through speeches and written submissions, and voting in plenary. 
%
The formal committee structure of the Parliament provides strategic advantages to certain MEPs by providing committee members with privileged access to information, and opportunity to shape the Parliament's negotiation stance. This has led MEPs to self-select into committees dealing with issues that they find salient in order to affect outcomes in those policy areas \cite{bowler1995organizing,yordanova2009rationale}. 

Committee chairs hold important administrative powers to set the committee agenda and the topics for debate at committee meetings. Rapporteurs are tasked with preparing reports about committee activities, and represent a medium for disseminating information about committee activities to the broader plenary \cite{benedetto2005rapporteurs}. Rapporteurs thus plays a central role in shaping the image of committee activities available to committee outsiders. 
%
%
Outside of committees, strict institutional rules also govern the allocation of speaking time during the Parliament's plenary sessions, and structure the ability of MEPs to intercede during negotiations \cite{hix2006dimensions,proksch2010position}. The total amount of speaking time for any particular issue is limited and divided between time reserved for actors with formal duties in plenary such as rapporteurs, and time proportionally divided between party groups based upon their overall share of MEPs elected. Limits on speaking time can lead to competition between MEPs, and party group leaders allocate the scare resource of speaking time between competing demands from rank and file MEPs for maximum impact. 


Due to the limits in the total amount of speaking time available, MEPs can also submit written questions and statements that are appended to the plenary records. These provide extra opportunity for MEPs to state their positions outside the time limits imposed on oral questioning during plenary debates. These written questions have been found to be the most popular avenue used by MEPs to interact with the Commission directly \cite{raunio1996parliamentary}, and provide the opportunity for `fire-alarm oversight' of national governments guilty of implementation failures of EU law \cite{Jensen:2013et}. MEPs enjoy more discretion over their ability to submit written submissions than they do over oral speaking time.

In terms of the content of legislative speeches in the Parliament, it has been shown that speeches reflect latent ideological conflict between MEPs, with both left-right and pro-/anti-EU integration dimensions of conflict having been detected \cite{Slapin:2010il}. Using text analysis techniques based upon word-frequency distributions, these authors were able to demonstrate the correspondence between the content of legislative speeches and other measures of ideological positions found in the literature based upon roll-call votes and expert surveys. 

\subsection{Topic Models}

In the field of data analytics advanced topic modeling algorithms that go beyond word-frequency distributions have recently been applied to large-scale text collections. 
Considerable research on topic modeling has focused on the use of probabilistic methods such as variants of Latent Dirichlet Allocation (LDA) \cite{steyvers06prob}. Authors have subsequently developed analogous probabilistic approaches for tracking the evolution of topics over time in a sequentially-organized corpus of documents, such as the dynamic topic model (DTM) of Blei and Lafferty \cite{blei06dynamic}.  Alternative algorithms, such as Non-negative Matrix Factorization (NMF) \cite{lee99nmf}, have also been effective in discovering the underlying topics in text corpora \cite{ocallaghan15eswa,wang12group}. Saha \& Sindhwani \cite{saha12learning} proposed an online learning framework for employing NMF to extract topics from streaming social media content, by dividing the streams into short sliding time windows so as discover topics that are smoothly evolving over time. 

As well as analyzing temporal data, recent work in this area has focused on important practical issues, including automating parameter selection (\eg how many topics are appropriate for our corpus?) and assessing \emph{topic coherence} (\ie how meaningful are the topics generated by our algorithm?) \cite{chang09tea,ocallaghan15eswa}. The latter corresponds closely to the concept of \emph{semantic validity} introduced in \cite{quinn2010analyze} for assessing the reliability of topics found in text corpora. This concept covers both intra-topic validity (the extent to which a single topic is meaningful) and inter-topic validity (the extent to which different topics are related to one another in a meaningful way).

\subsection{Topic Models Applied to Political Texts}

Some topic modeling methods have been adopted in the political science literature to analyze political attention. In settings where politicians have limited time-resources to express their views, such as the plenary sessions in parliaments, politicians must decide what topics to address. Analyzing such speeches can thus provide insight into the political priorities of the politician under consideration. Single membership topic models, that assume each speech relates to one topic, have successfully been applied to plenary speeches made in the 105th to the 108th U.S. Senate in order to trace political attention of the Senators within this context over time \cite{quinn2010analyze}. 
This study found that a rich and meaningful political agenda emerged from the collected speeches, where topics  evolved significantly over time in response to both internal and external stimuli. 


Bayesian hierarchical topic models have also been used to capture the political priorities of Members of Congress as found in their official press releases \cite{grimmer10bayesian}. This study shows that the press releases are also responsive to external stimuli such as upcoming votes in Congress or events external to Congress such as the anniversary of September 11th. Press release topics are also geographically structured with Members of Congress from rural farming communities more likely to pay attention to agricultural issues than those from urban communities for instance. The introduction of these methods to the study of political attention has allowed researchers to consider larger and more complete datasets of political activity across longer time periods than has previously been possible. The results unveil latent patterns in political attention that are difficult and time-consuming to capture using more traditional methodological approaches, such as expert surveys and hand-coding of texts. Applying them to study the political agenda of the European Parliament is the aim of this paper.


\section{Methods}
\label{sec:methods}

In this section we describe a two-layer strategy for applying topic modeling in a non-negative matrix factorization framework to a timestamped corpus of political speeches. Firstly, in \refsec{sec:methods1} we describe the application of NMF topic modeling to a single set of speeches from a fixed time period. 
Secondly, in \refsec{sec:methods2} we propose a new approach for combining the outputs of topic modeling from successive time periods to detect a set of \emph{dynamic topics} that span part or all of the duration of the corpus. 

\subsection{Topic Modeling Speeches}
\label{sec:methods1}
While work on topic models often involves the use of LDA, NMF can also be applied to textual data to reveal topical structures \cite{wang12group}. The ability of NMF to apply TF-IDF weighting to the data prior to topic modeling has shown to be advantageous in producing diverse but semantically coherent topics which are less likely to be represented by the same high frequency terms. This makes NMF suitable when the task is to identify both broad, high-level groups of documents and niche topics with specialized vocabularies \cite{ocallaghan15eswa}. 

Given a corpus of $n$ speeches, we first construct a speech-term frequency matrix $\m{A} \in \Real^{n\times m}$, where $m$ is the number unique terms present across all speeches (\ie the corpus vocabulary). Applying NMF to $\m{A}$ results in a reduced rank-$k$ approximation in the form of the product of two non-negative factors $\m{A} \approx \m{W}\m{H}$, where the objective is to minimize the reconstruction error between $\m{A}$ and $\m{W}\m{H}$. The rows of the factor $\m{H} \in \Real^{k\times m}$ can be interpreted as $k$ topics, defined by non-negative weights for each of the $m$ terms in the corpus vocabulary. Ordering each row provides a topic descriptor, in the form of a ranking of the  terms relative to corresponding topic. The columns in the matrix $\m{W} \in \Real^{n\times k}$ provide membership weights for all $n$ speeches with respect to the $k$ topics. 

In our experiments we use the fast alternating least squares variant of NMF introduced in \cite{lin07gradient}. NMF algorithms are often initialized with random  factors. However, this can lead to unstable results, where the algorithm converges to a variety of different local minima of poor quality, depending on the random initialization. To ensure a deterministic output and to improve the quality of the resulting topics, we generate initial factors using the Non-negative Double Singular Value Decomposition (NNDSVD) approach \cite{bout08headstart}.

\begin{figure}[!t]
    \centering
    \includegraphics[width=3.1in]{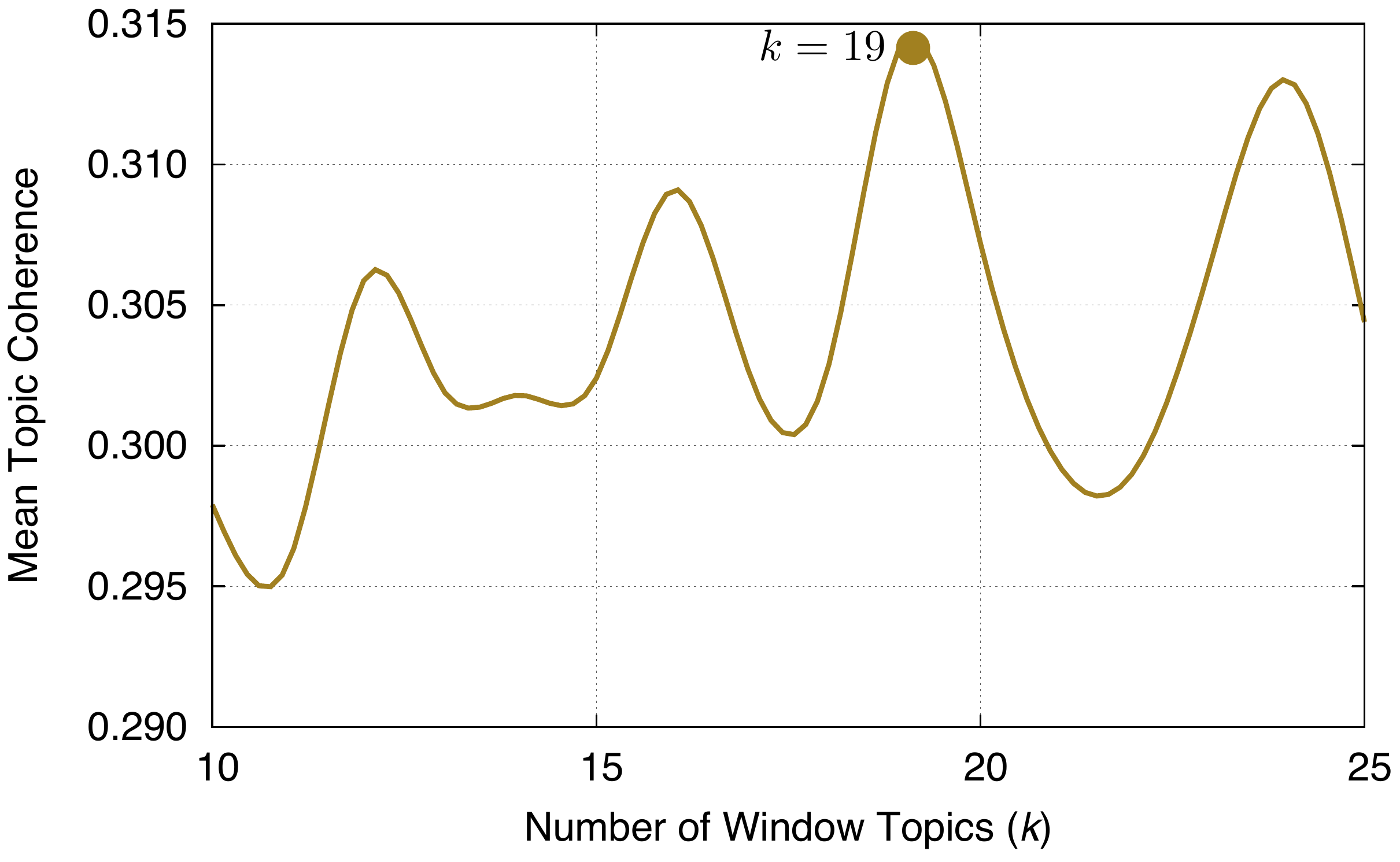}
    \vskip -0.9em
	\caption{Plot of mean TC-W2V topic coherence scores for different values for the number window topics $k$, generated on a time window of European Parliament speeches from 2005-Q1.}
    \label{fig:windowk}
    \vskip -0.5em
\end{figure}

A key parameter selection decision in topic modeling pertains to the number of topics $k$. Choosing too few topics will produce results that are overly broad, while choosing too many will lead to many small, highly-similar topics. One general strategy proposed in the literature has been to compare the \emph{topic coherence} of topic models generated for different values of $k$ \cite{chang09tea}. A range of such coherence measures exist in the literature, although many of these are specific to LDA. Recently, O'Callaghan \etal \cite{ocallaghan15eswa} proposed a  general measure, TC-W2V, which evaluates the relatedness of a set of top terms describing a topic, based on the similarity of their representations in a \emph{word2vec} distributional semantic space \cite{mikolovEfficient}. Specifically, the coherence of a topic $t_h$ represented by its $t$ top ranked terms is given by the mean pairwise cosine similarity between all relevant term vectors in the \emph{word2vec} space:
\begin{equation}
\textrm{coh}(t_h) = \frac{1}{\binom{t}{2}}\sum_{j=2}^{t} \sum_{i=1}^{j-1} cos(wv_i, wv_j)
\end{equation} 
An overall score for a topic model $T$ consisting of $k$ topics is given by the mean of the individual topic coherence scores:
\begin{equation}
\textrm{coh}(T) =  \frac{1}{k} \sum_{h=1}^{k}\textrm{coh}(t_h) 
\label{eqn:meancoh}
\end{equation} 
An appropriate value for $k$ can be identified by examining a plot of the mean TC-W2V coherence scores for a fixed range [$k_{min},k_{max}$] and selecting a value corresponding to the maximum coherence. An example is shown in \reffig{fig:windowk}, where the plot of mean coherence scores suggests a value $k=19$ from a candidate range $[10,25]$.

Parliamentary speeches will often be short and concise. In the case of the EP, speeches are often limited to 1-2 minutes in duration. As such, we would expect each speech to be primarily related to a single topic. This is consistent with the observations made by Quinn~\etal \cite{quinn2010analyze} when analyzing speeches from the U.S. Congress. Here we produce a single membership topic model (\ie a disjoint clustering of individual speeches in relation to topics) by selecting the maximum membership weight for each row in the factor $\m{W}$. 


	
\subsection{Dynamic Topic Modeling}
\label{sec:methods2}

When applying clustering to temporal data, authors have often proposed dividing the data into \emph{time windows} of fixed duration \cite{sulo10temporal}. In the case of streaming data, such as content originating from social media platforms, this involves artificially transforming the continuous streams into sliding windows. However, in the case of political speeches transcribed from distinct plenary sessions, the data is naturally divided into segments. While some aspects of the agenda will remain common between successive sessions, in other cases the focus of debates will change considerably between sessions. Consequently, online learning approaches which use sliding windows and assume a smooth evolution in topics over time, such as proposed in \cite{saha12learning}, may be unsuitable. 

Following \cite{sulo10temporal}, we divide the full time-stamped corpus of parliamentary speeches into $\tau$ disjoint time windows $\fullset{W}{\tau}$ of equal length. 
The rationale for the use of time windows as opposed to processing the full corpus in batch is two-fold: 1) we are interested in identifying the agenda of the parliament at individual time points as well as over all time; 2) short-lived topics, appearing only in a small number of time windows, may be obscured by only analyzing the corpus in its entirety.

At each time window $W_i$, we apply NMF with parameter selection based on \reft{eqn:meancoh} to the transcriptions of all speeches delivered during that window, yielding a \emph{window topic model} $T_{i}$ containing $k_{i}$ \emph{window topics}. This process produces a set of successive window topic models $\fullset{T}{\tau}$, which represents the output of the first layer in our proposed methodology.

From the window topic models we construct a new condensed representation of the original corpus, by viewing the rows of each factor $\m{H}_{i}$ coming from each window topic model as ``topic documents''. Each topic document naturally contains non-negative weights indicating the descriptive terms for that window topic. We expect that window topics from different windows which share a common theme will have similar topic documents. Specifically, we construct a condensed topic-term matrix $\m{B}$ as follows:
\begin{enumerate}
\item Start with an empty matrix $\m{B}$.
\item For each window topic model $T_{i}$:
\begin{enumerate}
\item For each window topic within $T_{i}$, select the $t$ top ranked terms from the corresponding row vector of the associated NMF factor $\m{H}$, set all weights for all other terms in that vector to 0. Add the vector as a new row in $\m{B}$.
\end{enumerate}
\item Once vectors from all topic models have been stacked in this way, remove any columns with only zero values (\ie terms from the original corpus which did not ever appear in the $t$ top ranked terms for any window topics).
\end{enumerate}
The matrix $\m{B}$ has size $n' \times m'$, where $n' = \sum_{i=1}^{\tau} k_{i}$ is the total number of ``topic documents'' and $m' << m$ is is the subset of relevant terms remaining after Step 3. The use of only the top $t$ terms in each topic document allows us to implicitly incorporate feature selection into the process. The result is that we include those terms that were highly-descriptive in each time window, while excluding those terms that never featured prominently in any window topic. This reduces the computational cost for the second factorization procedure described below.

Having constructed $\m{B}$, we now apply a second layer of NMF topic modeling to the matrix to identify $k'$ \emph{dynamic topics} which potentially span multiple time windows. The process is the same as that outlined previously in \refsec{sec:methods1}. Here the TC-W2V coherence measure is used to detect number of dynamic topics $k'$. The resulting factors can be interpreted as follows: the top ranked terms in each row of $\m{H}$ provide a description of the dynamic topics; the values in the columns of $\m{W}$ indicate to what extent each window topic is related to each dynamic topic.

We track the evolution of these topics over time as follows. Firstly, we assign each window topic to the dynamic topic for which it has the maximum weight, based on the values in each row in the factor $\m{W}$. We define the temporal \emph{frequency} of a dynamic topic as the number of distinct time windows in which that dynamic topic appears. The set of all speeches related to this dynamic topic across the entire corpus corresponds to the union of the speeches assigned to the individual time window topics which are in turn assigned to the dynamic topic. 
%
To summarize, the key outputs of the two-layer topic modeling process are as follows:
\begin{enumerate}
\item A set of $\tau$ topic models, one per time window, each containing $k_{i}$ \emph{window topics}. These are described using their top $t$ terms and the set of all associated speeches.
\item A set of $k'$ \emph{dynamic topics}, each with an associated set of window topics. These are described using their top $t$ terms and set of all associated speeches.
\end{enumerate}
\reftab{fig:eg-dynamic} shows a partial example of a dynamic topic. We observe that, for the four window topics, there is a common theme pertaining to climate change. While the variation across the term lists reflects the evolution of this dynamic topic over the corresponding time period (2008-Q4 to 2010-Q1), the considerable number of terms shared between the lists underlines its semantic validity.

\begin{table}[!t]
\centering
\scriptsize{
\begin{tabular}{|c|p{1.3cm}|p{1.3cm}|p{1.3cm}|p{1.3cm}|}
\hline\textbf{Rank} & \textbf{2008-Q4} & \textbf{2009-Q1} & \textbf{2009-Q4} & \textbf{2010-Q1} \\\hline
1             & energy           & climate          & climate       & climate    \\
2             & climate          & change           & change        & copenhagen   \\
3             & emission         & future           & copenhagen    & change   \\
4             & package          & emission         & developing    & summit   \\
5             & change           & integrated       & emission      & emission   \\
6             & renewable        & water            & conference    & international   \\
7             & target           & policy           & summit        & mexico   \\
8             & industry         & target           & agreement     & conference   \\
9             & carbon           & industrial       & global         & global  \\
10            & gas              & global           & energy        & world  \\\hline
\end{tabular}
}
\caption{Example of 4 window topics, described by lists of top 10 terms, which have been grouped together in a single dynamic topic related to climate change.}
\label{fig:eg-dynamic}
\end{table}



\section{Data}
\label{sec:data}

During August 2014 we retrieved all plenary speeches available on Europarl, the official website of the European Parliament\footnote{\url{http://europarl.europa.eu}}, corresponding to parliamentary activities of MEPs during the 5th -- 7th terms of the EP. This resulted in 269,696 unique speeches in 24 languages. While we considered the use of either multi-lingual topic modeling or automated translation of documents, issues with the accuracy and reliability of both strategies lead us to focus on English language speeches in plenary -- either from native speakers or translated -- which make up the majority of the speeches available on Europarl. A corpus of 210,247 English language speeches was identified in total, representing 77.95\% of the original collection. In terms of coverage of speeches from MEPs from the member states, this ranged from 100\% for the United Kingdom, through 87\% for Germany, down to 66.2\% for Romania. However, the most recent state to accede to the EU, Croatia, represents an outlier in the sense that only 2.6\% of speeches were available in English at the time of retrieval due to EP speech translation issues.

We subsequently divided the corpus into 60 quarterly time windows, from 1999-Q3 to 2014-Q2. We selected a quarter as the time window duration to allow for the identification of granular topics, while also ensuring there there existed a sufficient number of speeches in each time window to perform meaningful topic modeling. In particular, we wished to avoid empty time windows occurring due to the summer recess of the EP. For each time window $W_{i}$ we construct a speech-term matrix $\m{A}_{i}$ as follows:
\begin{enumerate}
\item Select all speech transcriptions from window $W_{i}$, and remove all header and footer lines.
\item Find all unigram tokens in each speech, through standard case conversion, tokenization, and lemmitization.
\item Remove short tokens with < 3 characters, and tokens corresponding to generic stop words (\eg ``are'', ``the''), parliamentary-specific stop words (\eg ``adjourn'', ``comment'') and names of politicians as listed on the EP website.
\item Remove tokens occurring in < 5 speeches.
\item Construct $\m{A}_{t}$, based on the remaining tokens. Apply standard TF-IDF term weighting and document length normalization.
\end{enumerate}
The resulting time window data sets range in size from 679 speeches in 2004-Q3 to 9,151 speeches in 2011-Q4, with an average of 4,811 terms per data set. 

\section{Experimental Results}
\label{sec:eval}

\subsection{Experimental Setup}
\begin{figure}[!t]
    \centering
    \includegraphics[width=3.05in]{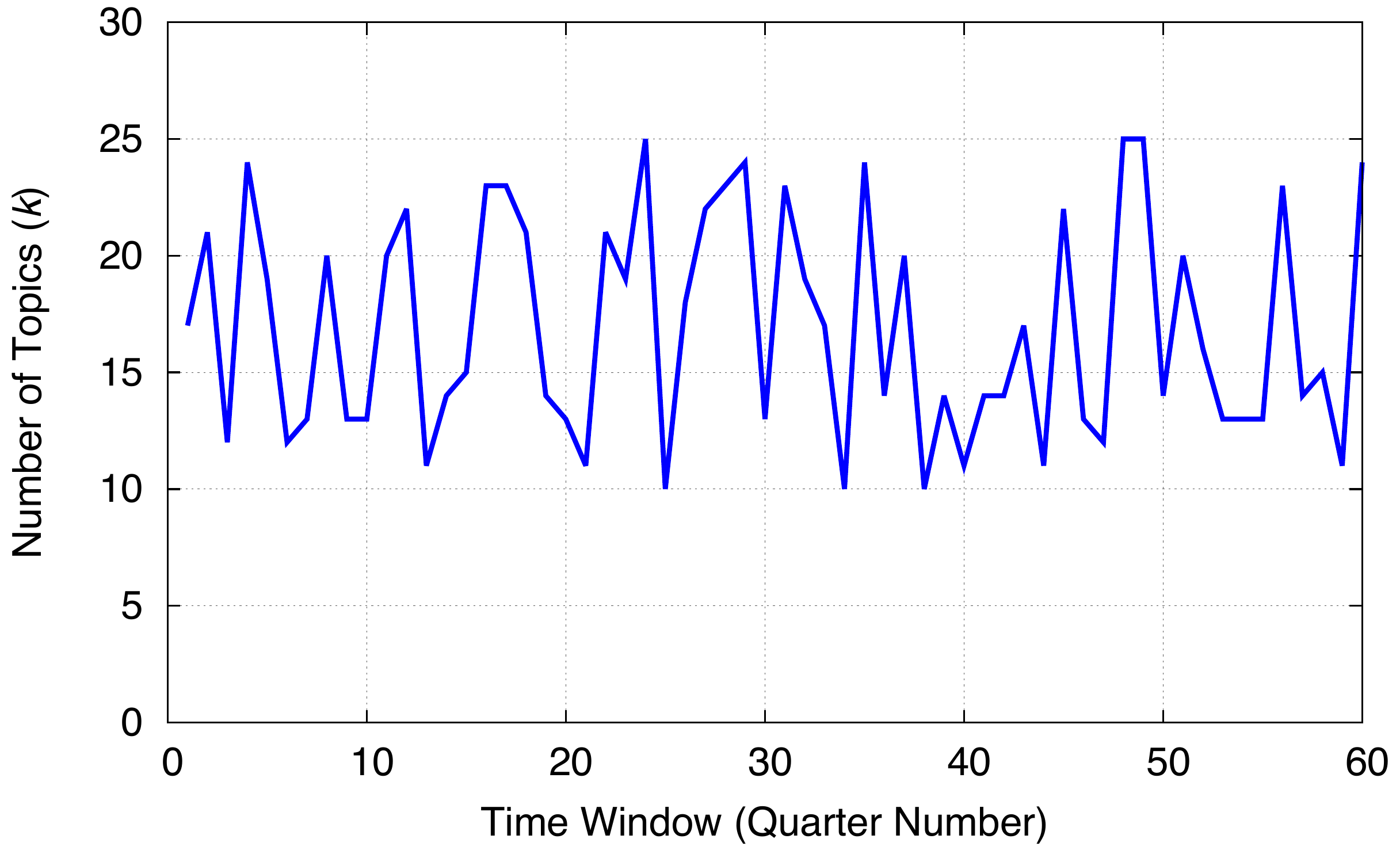}
    \vskip -1em
	\caption{Number of window topics identified per time window, from 1999-Q3 (\#1) to 2014-Q2 (\#60).}
    \label{fig:numtopics}
\end{figure}

After pre-processing the data, to identify window topics we applied NMF with parameter selection as described in \refsec{sec:methods1}. Given the relatively specialized vocabulary used in EP debates, when building the \emph{word2vec} space for parameter selection, as our background corpus we used the complete set of English language speeches. We used the same \emph{word2vec} settings and number of top terms per topic ($t=10$) as described in \cite{ocallaghan15eswa}. At each time window, we generated models containing $k \in [10,25]$ window topics, selecting the value $k$ that maximized mean TC-W2V coherence. The resulting median number of topics per window was 15.5. The illustration of the number of topics per window in \reffig{fig:numtopics} shows that there is considerable variation in the number of topics detected for each window, which does not correlate with the number of speeches per quarter (Pearson correlation $0.006$).

\begin{figure}[!t]
    \centering
    \includegraphics[width=3.05in]{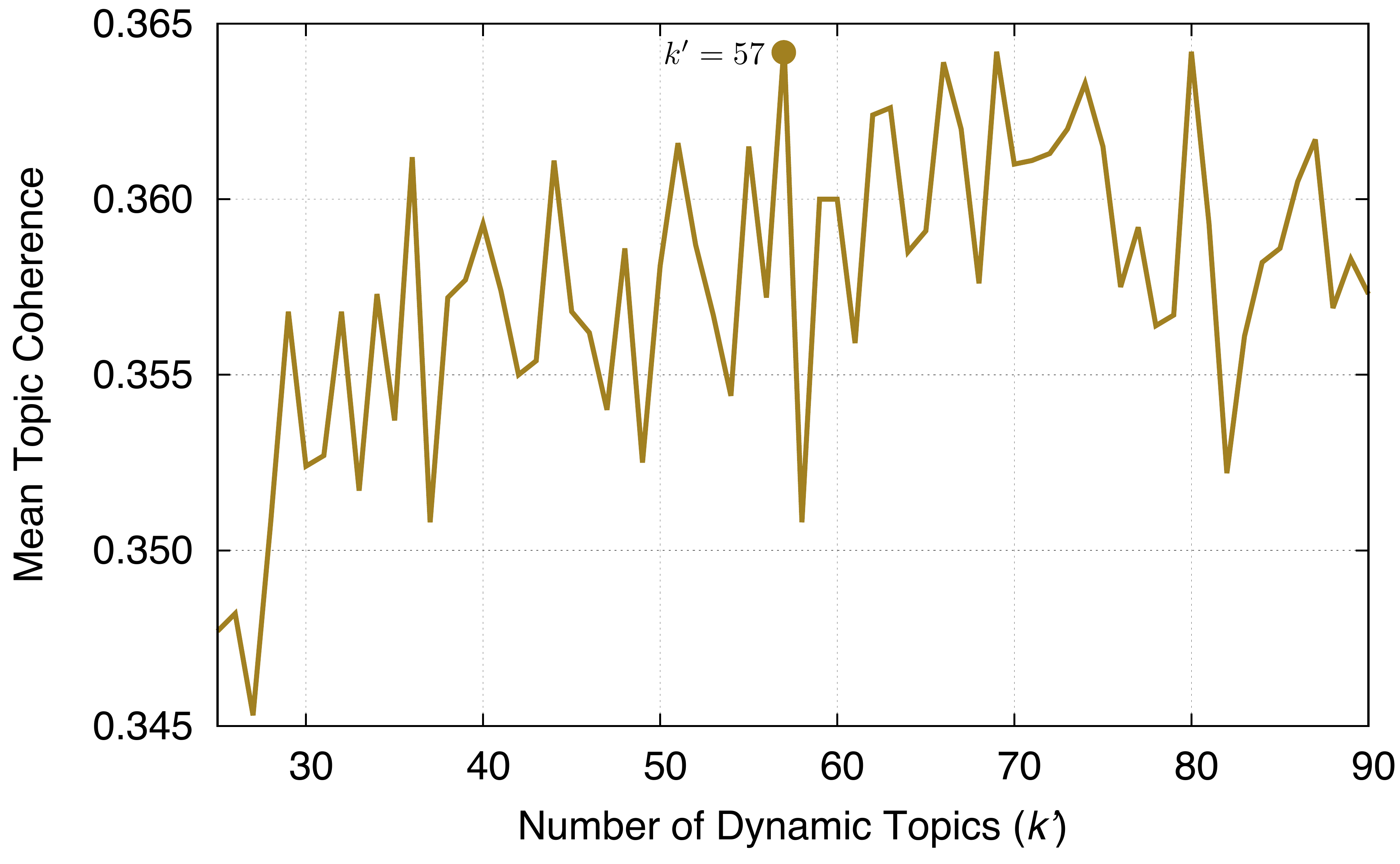}
    \vskip -1em
	\caption{Plot of mean TC-W2V topic coherence scores for different values for the number dynamic topics of $k'$, across a candidate range $[25,90]$.}
    \label{fig:numdynamic}
\end{figure}

The process above yielded 1,017 window topics across the 60 time window. We subsequently applied dynamic topic modeling as described in \refsec{sec:methods2}. For the number of terms $t$ representing each window topic, we experimented with values from 10 to the entire number of terms present in a time window. However, values $t > 20$ did not result in significantly different dynamic topics. Therefore, to minimize the dimensionality of the data, we selected $t=20$. This yielded a matrix of 1,017 window topics represented by 2,710 distinct terms. We applied parameter selection based on TC-W2V coherence to select an overall number of dynamic topics $k'$ from a range $k' \in [25,90]$. The resulting plot (see \reffig{fig:numdynamic}) indicated a maximal value at $k'=57$, although a number of close peaks exist in the range [62,80]. This corresponds to our manual inspections of the results, where the topic models for these values of $k$' appeared to be highly similar, with minor variations corresponding to merges or splits of strongly-related topics.

\subsection{Dynamic Topic Validation}
\label{sec:validation}

\begin{table*}[!t]
\centering
\scriptsize{
\begin{tabular}{|r|l|l|c|c|}
\hline
\textbf{Topic} & \textbf{Short Label}             & \textbf{Top 10 Terms}                                                                          & \textbf{Coh.} & \textbf{Freq.} \\\hline
13             & Transport                        & transport, railway, rail, passenger, road, network, freight, system, train, infrastructure              & 0.54          & 19             \\
42             & The Balkans                      & kosovo, serbia, balkan, resolution, bosnia, albania, iceland, herzegovina, macedonia, process           & 0.50          & 12             \\
33             & Air transport                    & air, passenger, transport, aviation, airport, traffic, airline, flight, sky, single                     & 0.48          & 10             \\
29             & Adjusting to globalisation       & fund, globalisation, egf, worker, adjustment, mobilisation, european, redundant, application, eur       & 0.47          & 15             \\
6              & Energy                           & energy, gas, renewable, efficiency, supply, source, electricity, market, target, project                & 0.47          & 36             \\
39             & Education and culture             & programme, education, culture, language, cultural, youth, sport, learning, young, training              & 0.43          & 21             \\
8              & Fisheries                        & fishery, fishing, fish, stock, fisherman, fleet, sea, common, policy, measure                           & 0.43          & 34             \\
2              & Human rights                     & rights, human, fundamental, freedom, democracy, law, charter, resolution, union, violation              & 0.43          & 52             \\
45             & Maritime issues                  & port, sea, maritime, safety, ship, accident, oil, vessel, transport, inspection                         & 0.43          & 10             \\
21             & Healthcare                       & health, patient, environment, safety, public, care, healthcare, action, disease, mental                 & 0.42          & 18             \\
26             & Child protection                 & child, internet, pornography, sexual, school, exploitation, young, victim, education, crime             & 0.42          & 14             \\
56             & Road safety                      & road, safety, vehicle, transport, system, driver, accident, motor, noise, ecall                         & 0.41          & 12             \\
16             & Research                         & research, programme, innovation, framework, funding, industry, technology, development, cell, institute & 0.41          & 15             \\
15             & Turkish accession                & turkey, turkish, accession, progress, cyprus, negotiation, union, membership, croatia, macedonia        & 0.41          & 20             \\
35             & Tax                              & tax, vat, taxation, rate, system, fraud, states, evasion, car, transaction                              & 0.41          & 11             \\
32             & Trade - WTO and aid            & trade, wto, world, development, developing, international, negotiation, aid, free, relation             & 0.39          & 19             \\
47             & Product labelling and regulation & product, medicinal, medicine, tobacco, labelling, safety, consumer, regulation, organic, advertising    & 0.39          & 11             \\
11             & Trade - Trade partnerships     & agreement, partnership, morocco, trade, negotiation, data, cooperation, association, korea, fishery     & 0.39          & 18             \\
49             & Regional funds                   & policy, region, cohesion, development, regional, strategy, structural, fund, economic, area             & 0.39          & 22             \\
17             & CFSP                             & security, policy, defence, common, foreign, military, nato, immigration, aspect, european               & 0.39          & 19            \\\hline
\end{tabular}
}
\caption{List of top 20 dynamic topics, ranked by their TC-W2V topic coherence. For each dynamic topic, we report a manually-assigned short label, the top 10 terms, coherence, and frequency (\ie number of time windows in which it appeared).}
\label{tab:topics}
\end{table*}

The 57 topics identified in our experiments are diverse, both in terms of their thematic content and temporal signatures. \reftab{tab:topics} lists the top 20 dynamic topics in the data, ranked with respect to their TC-W2V topic coherence scores. We report the temporal frequency of the topics, together with a manually-assigned short label for discussion purposes\footnote{Full details of all window topics and dynamic topics are available at \url{http://erdos.ucd.ie/europarl}}. The frequency of dynamic topics ranged from 11 which appeared in $< 10$ time windows, to a broad `Plenary administration' topic which appeared in 57 out of 60 windows. 

In general, we observed two distinct categories of dynamic topics. The first reflects the day-to-day politics of EU in terms of legislating and debating issues related to the core EU competencies (\eg `Energy', `Agriculture'), while the other reflects unanticipated exogenous shocks and MEPs reactions to these events (\eg Euro-crisis, September 11th attacks). These two categories exhibit differing temporal signatures. For instance, we see a considerable difference between the broad topic on fisheries policy (\reffig{fig:dynamic1}), when compared to the two topics arising from the events during the financial crisis and subsequent Euro-crisis as shown in Figures \ref{fig:dynamic2} and \ref{fig:dynamic3} respectively. This distinction between dynamic topic types reflects two different forms of political process in the Parliament. 

\begin{figure}[!t]
    \centering
    \includegraphics[width=3.08in]{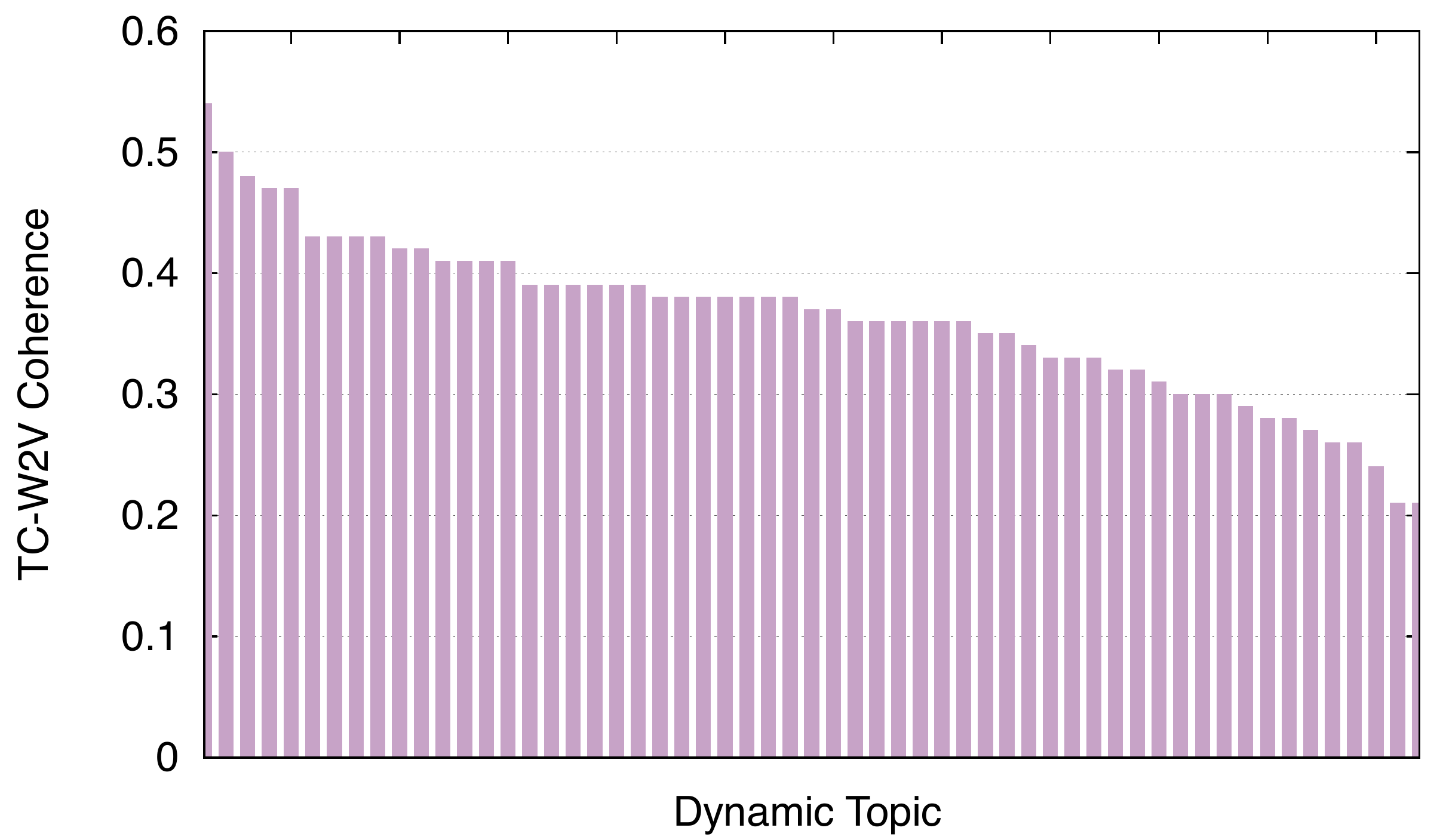}
    \vskip -0.6em
    \caption{Distribution of TC-W2V topic coherence values for 57 dynamic topics, based on top 10 terms.}
    \label{fig:coh}
\end{figure}

To examine the intra-topic semantic validity of these dynamic topics, \reffig{fig:coh} illustrates the distribution of TC-W2V coherence values for all dynamic topics, when evaluated in the \emph{word2vec} space built from the complete speech corpus. As evidence by the ranking in \reftab{tab:topics}, the most coherent topics often correspond to core EU competencies. Unsurprisingly, broad administrative topics prove to be least coherent (\eg `Commission questions', `Council Presidency', `Plenary administration'). Overall the mean topic coherence score of 0.36 is considerably higher than the lower bound for TC-W2V (-1.0), suggesting a high level of semantic validity across the board.

\begin{figure}[!t]
    \centering
    \includegraphics[width=3.1in]{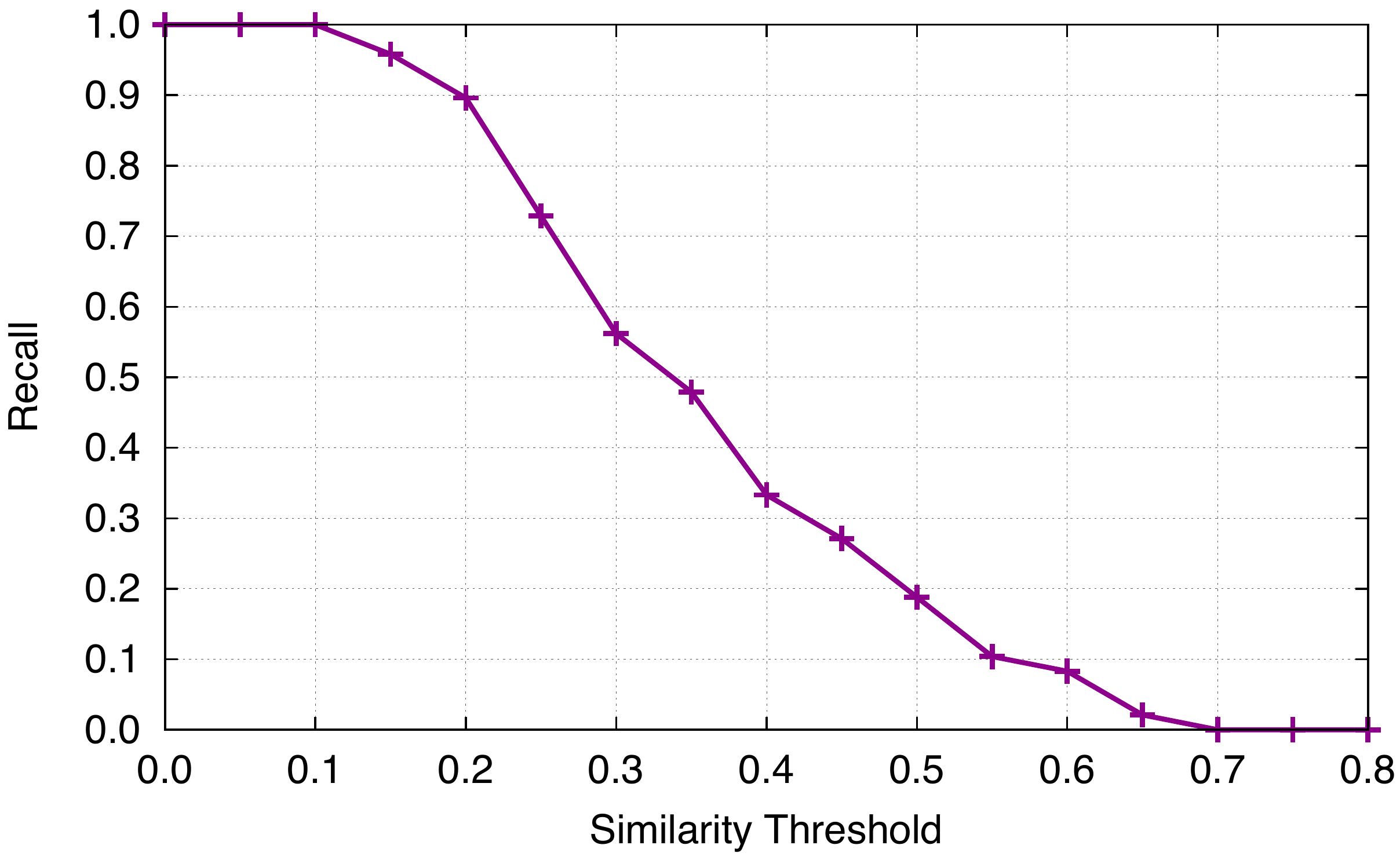}
    \vskip -1em
	\caption{Recall plot for EP taxonomy subjects relative to dynamic topics, for increasing thresholds for cosine similarity.}
    \label{fig:subjects}
\end{figure}

\begin{figure*}[!t]
    \centering
    \includegraphics[width=0.93\linewidth]{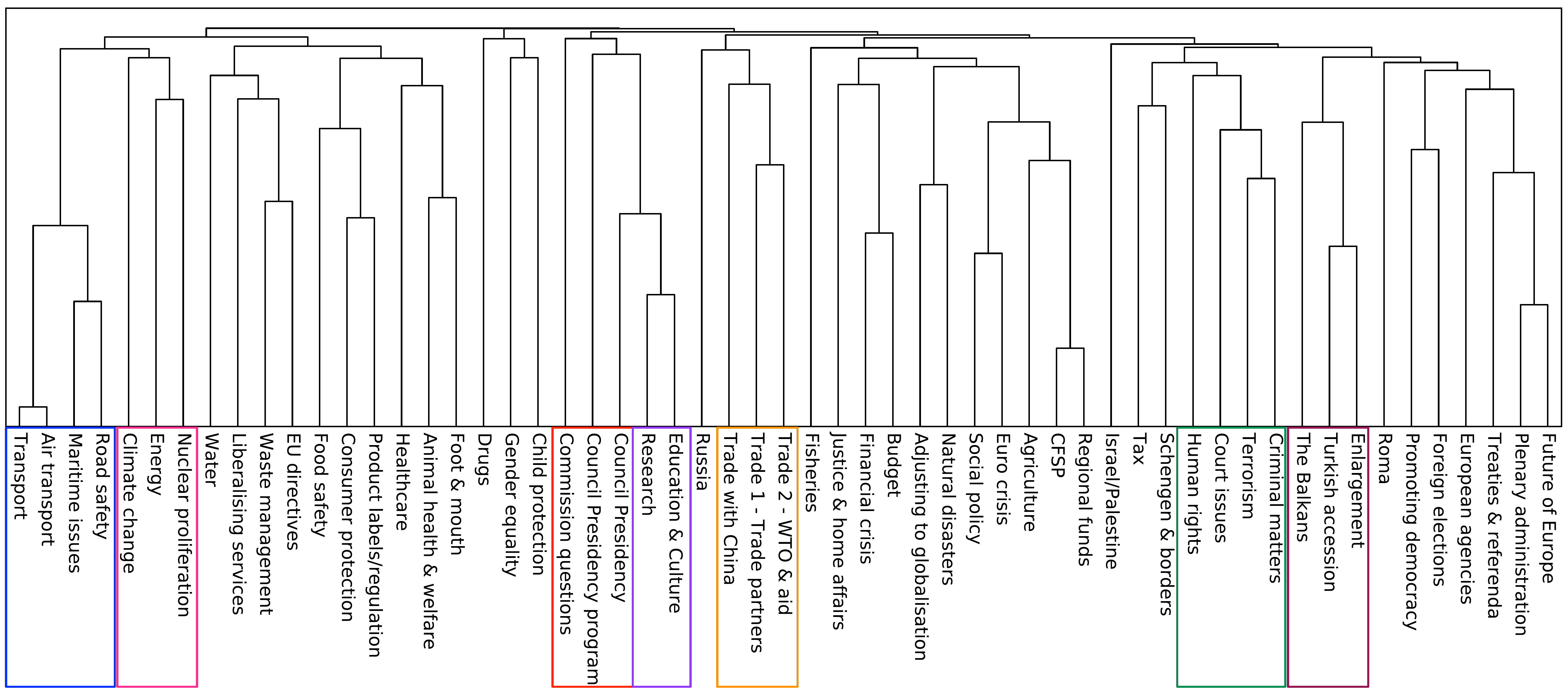}
    \vskip -1.2em
    \caption{Dendrogram for average linkage hierarchical agglomerative clustering of 57 dynamic topics.}
    \label{fig:tree}
\end{figure*}
\begin{table*}[!t]
\centering
\scriptsize{
\begin{tabular}{|l|l|c|}
\hline
\textbf{Subject}                                     & \textbf{Matched Topic: Top 10 Terms}                                                                           & \textbf{Sim.} \\\hline
1.10\;\;Fundamental Rights In The Union                 & rights, human, fundamental, freedom, democracy, law, charter, resolution, union, violation                    & 0.66          \\
4.40\;\;Education, Vocational Training \& Youth                 & programme, education, culture, language, cultural, youth, sport, learning, young, training                    & 0.63          \\
5.20\;\;Monetary Union                                           & euro, economic, growth, stability, pact, bank, policy, monetary, economy, ecb                                 & 0.62          \\
4.70\;\;Regional Policy                                          & policy, region, cohesion, development, regional, strategy, structural, fund, economic, area                   & 0.62          \\
3.50\;\;Research \& Technological Development               & research, programme, innovation, framework, funding, industry, technology, development, cell, institute       & 0.57          \\
3.60\;\;Energy Policy                                            & energy, gas, renewable, efficiency, supply, source, electricity, market, target, project                      & 0.53          \\
6.10\;\;Common Foreign \& Security Policy               & security, policy, defence, common, foreign, military, nato, immigration, aspect, european                     & 0.52          \\
3.20\;\;Transport Policy in General                              & transport, railway, rail, passenger, road, network, freight, system, train, infrastructure                    & 0.51          \\
4.60\;\;Consumers' Protection in General                         & product, medicinal, medicine, tobacco, labelling, safety, consumer, regulation, organic, advertising          & 0.50          \\
3.70\;\;Environmental Policy                                     & waste, recycling, directive, packaging, management, environment, electronic, fuel, environmental, radioactive & 0.50          \\\hline
\end{tabular}
}
\caption{Top 10 legislative procedure subjects with corresponding matching dynamic topics, ranked by cosine similarity of the match.}
\label{tab:subjects}
\end{table*}

To assess the inter-topic semantic validity of the results, we examine the extent to which any meaningful higher-level grouping exists among the 57 dynamic topics. To do this we apply average linkage agglomerative clustering to the topics. Following the approach described in \cite{greene08ensemble}, we re-cluster the row vectors from the second-layer NMF  factor $\m{H}$ using normalized Pearson correlation as a similarity metric. Here the vectors correspond the weights of each dynamic topic with respect to the 2,710 terms noted above. The dendrogram for the hierarchical clustering is shown in \reffig{fig:tree}. Using the interpretation provided in \cite{quinn2010analyze}, the lower the height at which any two  topics are connected in the dendrogram, the more similar their corresponding term usage patterns in EP sessions. 
We observe a number of higher-level groupings of interest, which are highlighted in \reffig{fig:tree}. These include groups related to transport in general, energy concerns, interactions with other institutions, education and research, trade relations, and EU enlargement. The presence of these higher-level associations between topics provide semantic validity for the results presented, where topics one might expect to be related are found to be correlated with respect to their rows in the NMF factor $\m{H}$ (\ie they share similar terms in their topic descriptors).


To externally quantify the extent to which the identified dynamic topics correspond to policy areas in which the EU has competencies, and thus provide evidence of construct validity, we compare the 57 dynamic topics to an existing taxonomy of subjects, which is used by Europarl to classify legislative procedures. The taxonomy as retrieved from the site has several different levels, ranging from broad top-level subjects (\eg `3 Community policies'), to highly-specific low-level subjects (\eg `3.10.06.05 Textile plants, cotton'). We compare our results to the second level of the taxonomy, containing 48 subjects (\eg `3.10 Agricultural policy and economies', `3.20 Transport policy in general'). For each subject code, we create a ``subject document'' consisting of the description of the subject and all lower level subjects within that branch of the taxonomy. We then identify the most similar dynamic topic by comparing the top 10 terms for that topic with each subject document, based on cosine similarity. \reftab{tab:subjects} shows the best matching subjects and topics identified using this approach. \reffig{fig:subjects} shows the \emph{recall} of all 48 subjects, for different threshold levels of cosine similarity. For instance, at a threshold of 0.25, suitably matching dynamic topics for $72.9\%$ of subjects are identified. To give a couple of examples, the topic hand-coded as relating to `Tax' from our topic model was correctly matched with the Europarl subject code `2.70 Taxation' broadly defined at level-2 of the taxonomy, and `2.70.01 Direct taxation' and `2.70.02 Indirect taxation' separately at level-3 of the taxonomy. When looking at the topic manually labeled as relating to `Drugs', cosine similarity matches this with the level-2 subject `4.20 Public health', which has a level-3 sub-category relating to `4.20.04 Pharmaceutical products and industry'. When taken in the context of the matches shown in \reftab{tab:subjects}, this indicates that our dynamic topics provide good coverage of the policy areas that might be expected to feature during EP debates, and thus increases our confidence in the construct validity of the method.


\subsection{Case Studies}
\label{sec:case}

In order to further investigate the construct validity of our topics, we focus on three specific examples that demonstrate how our modeling strategy captures variation in MEP attention to a topic over time, and how this attention is impacted upon by external stimuli. 

\begin{figure}[!t]
\centering
\subfigure[``Financial \& Euro crises'' dynamic topics]{\includegraphics[width=3.15in]{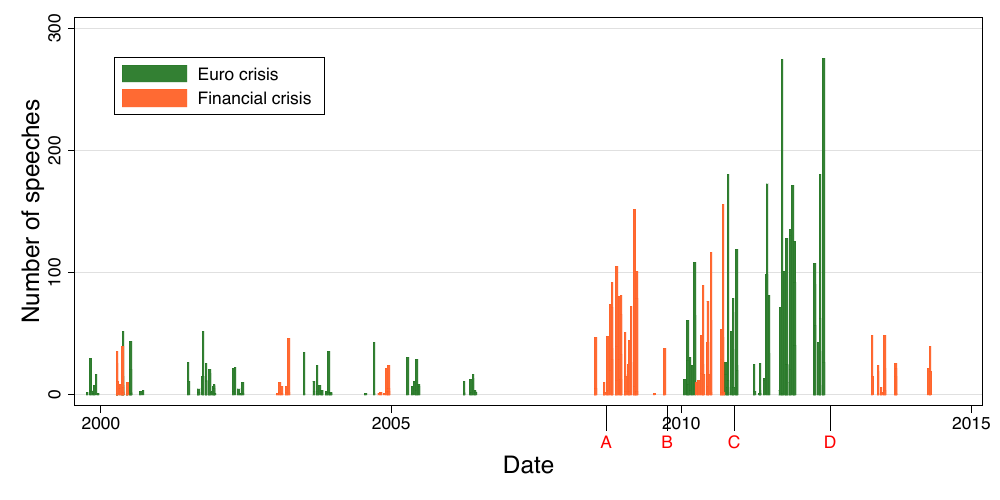}\label{fig:dynamic1}}
\subfigure[``Treaty changes \& referenda'' dynamic topic]{\includegraphics[width=3.15in]{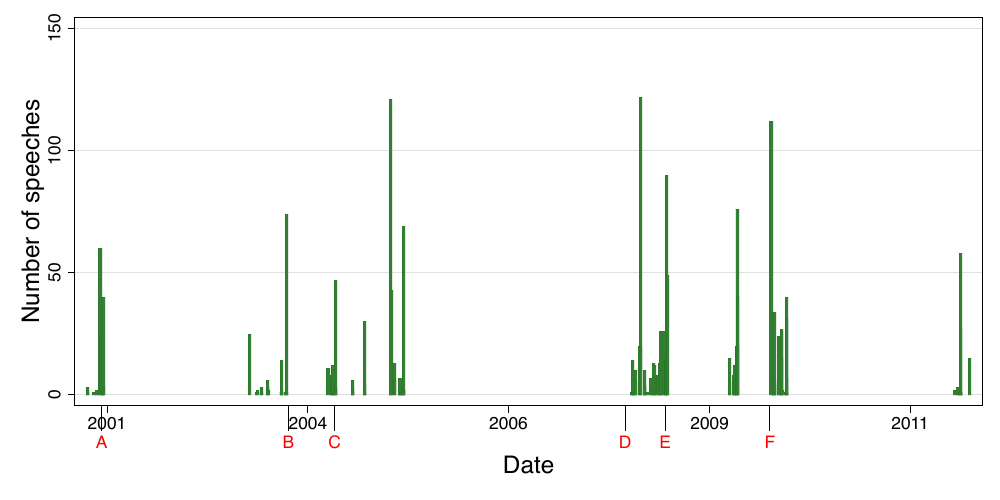}\label{fig:dynamic2}}
\subfigure[``Fisheries'' dynamic topic]{\includegraphics[width=3.15in]{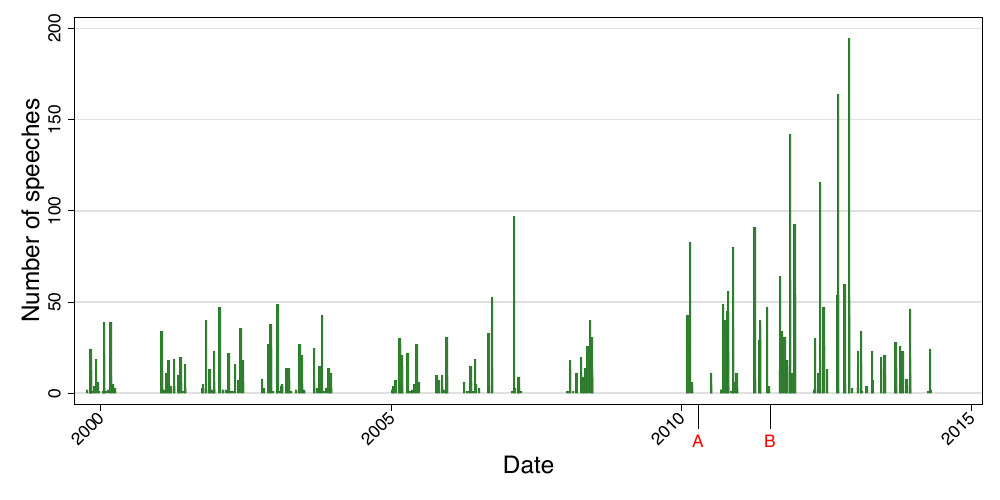}\label{fig:dynamic3}}
    \vskip -0.8em
\caption{Time plots for three sample dynamic topics across all time windows, from 1999-Q3 (\#1) to 2014-Q2 (\#60). Dates on the x-axis correspond to the dates on which speeches were made at EP plenary sessions.}
\label{fig:dynamic}
\vskip -0.8em
\end {figure}

Our first case study relates to MEP attention to the financial/Euro-crisis. The temporal distribution of speeches relating to this topic is illustrated in \reffig{fig:dynamic1}. This is an interesting case study, as the initial financial crisis peaked in 2008, and the Euro-crisis that followed has gone through a number of phases with major events in 2009, 2010 and 2012. As such, these events can be thought of as exogenous shocks that only garner MEP attention after they occur, and their exogenous nature provides a way to externally validity the dynamic topic modeling approach in use here. \reffig{fig:dynamic1} demonstrates a number of distinct peaks in MEP speech making on both the financial crisis topic (in orange) and the Euro-crisis topic (in green). Attention to the financial crisis starts to rise in 2008-Q3 and initially peaks in 2008-Q4 (point A in \reffig{fig:dynamic1}). This peak in activity corresponds to the date when the Lehman Brothers investment bank collapsed (15/9/2008). The other peaks in activity in \reffig{fig:dynamic1} correspond to important events in the Euro-crisis. Point B corresponds to the revelations about under-reporting of Greek debt following the Greek parliamentary election in October 2010, Point C to the Irish bailout in November 2010, and Point D to Mario Draghi's statement that the ECB was ``ready to do whatever it takes to preserve the euro" in the July 2012 respectively.

Our second case study relates to the process of EU treaty reform. This topic is of interest, because one would expect a large amount in variation in MEP attention to the topic over time, as Treaty revision and reform and the referenda that accompany them are rare event and should only garner MEP attention when such events occur. \reffig{fig:dynamic2} shows MEP attention to the treaty change and referenda topic between 2000 and 2014 in terms of the number of speeches associated with this topic. Three distinct treaties were discussed and debated over this period. The first was the Nice treaty, which was agreed upon in 2001 and put to the vote in a referendum in Ireland in June 2001. The `No' vote in that resulted from this referendum accounts for Point A in \reffig{fig:dynamic2}. The next set of treaty related events to occur were the negotiations and failed ratification of the Constitutional Treaty between 2003 and 2005. This process accounts for Point B in \reffig{fig:dynamic2}, that correspond to the Intergovernmental Conference negotiating the treaty text that begun in October 2003. In the end the Constitutional Treaty was rejected by the French and Dutch in referenda in May/June 2005. Point C indicates the date of the signing of the Enlargement treaty in May 2004. The Lisbon treaty was negotiated to replace the failed Constitutional treaty, and we observe a significant peak in MEP speeches directly relating to the Lisbon treaty when it was signed (Point D), and when the first Irish referendum failed to ratify the treaty in June 2008 (Point E). A similar peak in MEP speeches relating to treaty reform corresponds to the second Irish referendum that eventually approved the Lisbon treaty in October 2009 (Point F).

Our third and final case study relates to fisheries policy. Fisheries is an interesting theme for the dynamic topic modeling approach to detect, because it is more associated with the day-to-day functioning of the EU as a regulator of the fisheries industry, when compared to more headline making policies and events like the economic crisis and treaty changes. \reffig{fig:dynamic3} demonstrates the prevalence of the fisheries topic over time. As can be seen, MEPs are seen to pay a reasonably stable level of attention to fisheries in terms of the numbers of speeches being made between 2000 and 2010. This trend is interrupted in 2010, when an increase in MEP attention to the fisheries topic is observed. This can be explained by the fact that in 2009 the European Commission launched a public consultation on reforming EU fisheries policy, the results of which were presented to the Parliament and Council in April 2010. The launch of this working document corresponds to a increase in the number of MEP speeches related to the fisheries topic as detected by the dynamic topic model (Point A). The peak in MEP speech making relating to this topic (Point B) corresponds with Commissioner Maria Damanaki introducing a set of legislative proposals designed to reform the common fisheries policy in a speech to the European Parliament in July 2011.

In general, the fact that the variation over time  that we observe in MEP attention to these case study topics appears to be driven by exogenous events provides a form of construct validity for our topic modeling approach. 

\subsection{Explaining MEP Speech Counts}
\label{sec:explain}

We now focus our attention on the 7th European Parliament which sat between 2009--2014. We focus on this term as a set of interesting covariates are available at the MEP level that can help us explain MEP contributions to a given topic. The dependent variable we seek to explain is the observed variation in the number of speeches each MEP makes on each of our identified dynamic topics. We employ a count-model framework suitable for analyzing count data \cite{cameron2013regression}. The first issue to note with the count variable under consideration is that there is a large number of zeros. This is due to the fact that, for many topics, a considerable number of MEPs are recorded as making no speeches. This is likely due to the data-generating process in the topic model from which our dependent variable emerges. As described in \refsec{sec:methods1}, we apply a single membership topic modeling approach where each speech is associated with one topic. This assumption is generally unproblematic, given the short amount of time allowed for speeches and the concentrated nature of the messages MEPs seek to communicate in them. However, any speeches that might contain multiple topics are only counted towards a single topic in the model. The result is that, in some cases, the ``true'' number of topics addressed by MEPs is under-represented and an inflated zero count is observed. In order to account for the inflated zero count, we model MEP speech-making as a two-stage process using a zero-inflated negative binomial regression model \cite{cameron2013regression}. A zero-inflated negative binomial model includes a Logit regression component to capture the binary process determining whether or not a MEP speaks on a topic, and a negative-binomial regression component that seeks to capture the count process determining the number of speeches made, given that a MEP has chosen to speak on a topic.


In order to explain the variation observed in our dependent variable, we include variables relating to MEP's ideology, voting behavior, and the institutional structures in which they find themselves embedded within. We account for the left-right ideological position of a MEP's national party (as a proxy for MEP ideology) using data from \cite{scully2012national}. Following \cite{proksch2014politics}, we also include a measure of how often MEPs vote against their party group in favor of their national party and vice versa. The idea behind including these variables is that MEPs rebelling against one party in favor of another will either try to explain such behavior in their speeches thus increasing the count, or hide their behavior by making no speeches, thus decreasing the count. These data were taken from an updated version of the \cite{hix2006dimensions} dataset provided by those authors. In order to capture an MEP's committee positions we include dummies for committee membership, chairs, and Rapporteurs in committees that are directly related to a given topic. Committees were manually matched with topics to achieve this. We control for whether or not an MEP serves in the Parliamentary leadership. Controls are also included for the total number of speeches made by an MEP and the percentage of MEP speeches that are available in English as these are liable to affect the observed MEP speech count. Finally, we also include dummy variables to control for an MEP's country of origin, EP party-group membership, and the topic on which they are speaking. All institutional and control variables were scraped from the legislative observatory of the European Parliament. 



\begin{figure}[!t]
	\includegraphics[width=\linewidth]{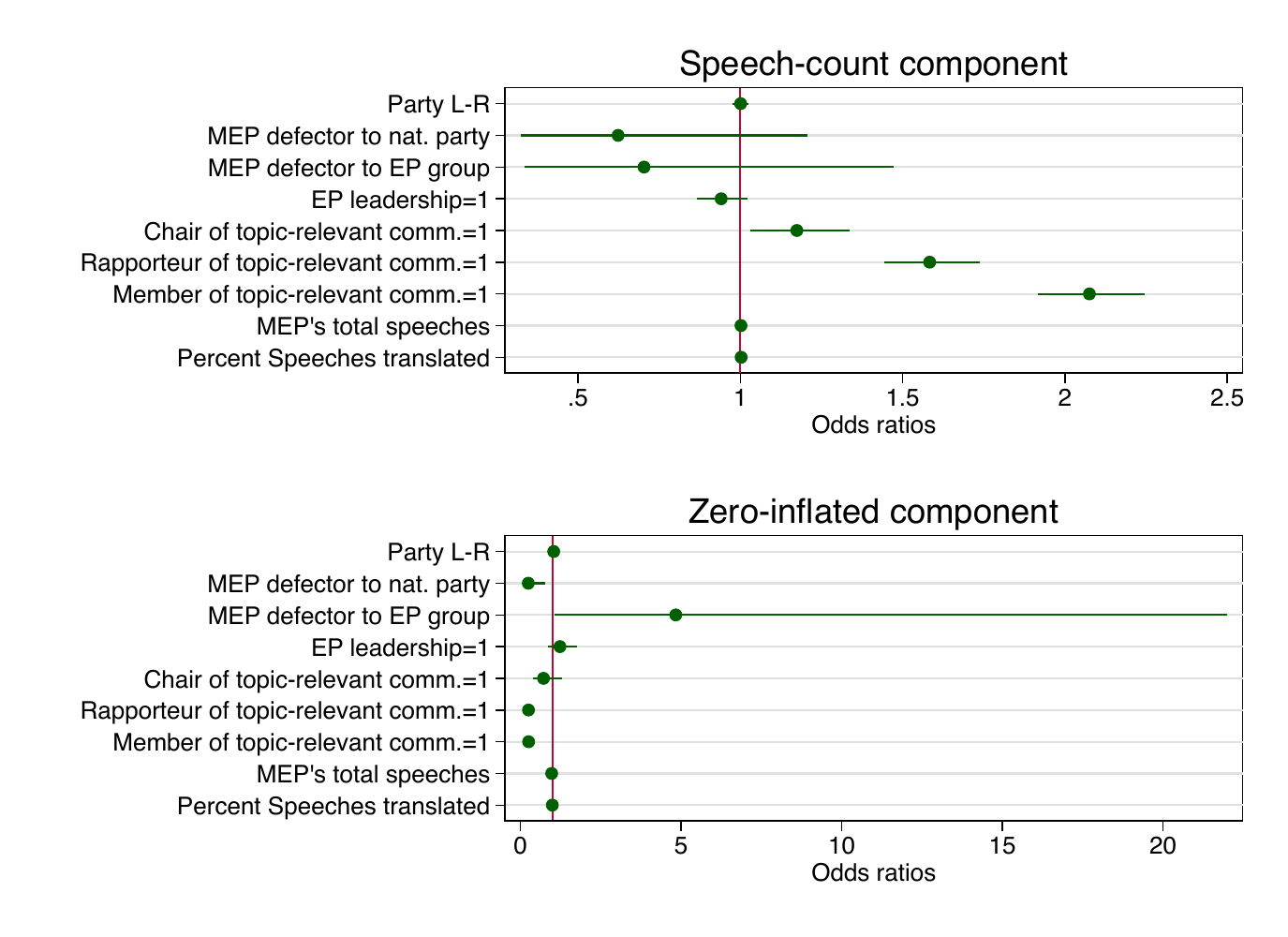}
	\vskip -0.5em
	\caption{Plot of coefficients for regression model.}
	\label{fig:regfig}
\end{figure}

The regression presented in \reffig{fig:regfig} provides further validation for the results of our topic modeling approach. The coefficients of the model have been exponentiated so as to represent odds ratios and aid interpretation. For the Logit component of the model accounting for zero inflation, an exponentiated coefficient value above 1 implies that an increase in that covariate leads to an increase in the odds that a zero is observed (no speech is made), while any value below 1 implies an increase that variable leads to a decrease in the odds of a zero being observed (a speech being made). For the count component of the model exponentiated coefficient values above 1 are interpreted as implying a positive relationship between the predictor and outcome variable, while values below 1 imply a negative relationship between the predictor and outcome variable.

We begin with the zero-inflated component of the model in \reffig{fig:regfig}. The model suggests that a MEP's national party ideology impacts upon whether or not they make speeches on a given topic, with more right-wing MEPs tending to make no topic speeches more often than left-wing MEPs. Furthermore, MEPs defecting to national parties tend to make speeches more often than those not defecting, while the opposite is true for MEPs defecting to EP groups from national parties. This is in line with the findings of \cite{proksch2014politics} who demonstrate that MEPs who rebel against their European party groups tend to make more speeches explaining why they do so, while those rebelling against their national party tend to make less speeches advertising their defection from the national party majority.

Of the institutionally related variables, holding a leadership position or a chair of a topic relevant committee has no significant relationship to MEP speech making, while being a member of a committee relevant to a topic, or holding a Rapporteurship for such a topic-relevant committee significantly impact upon whether or not MEPs make a speech that topic. The odds that a MEP makes no speeches on a given topic decrease by a factor of 0.255 if a MEP is a Rapporteur of a topic-relevant Committee and decrease by a factor of 0.259 if that MEP is a member of a topic-relevant committee.
The results also show that the odds of an MEP making no speeches on a given topic decrease for MEPs that make more speeches in total. The result relating to the percentage of MEP speeches that are in English (whether translated or originally so) is also found to be significant, suggesting that MEPs with more speeches available in English tend to make speeches on a given topic more often.


Moving to the speech-count component of the model, the results further reinforce our expectations that MEP positions within the Parliamentary committee system impact upon how much attention they pay to a particular topic. When an MEP holds a committee chair, Rapporteurship, or committee membership relevant to a particular topic, the odds that said MEP will make a speech on that topic increase by a factor of 1.173, 1.582, and 2.077 respectively.

\begin{figure}[!t]
    \centering
	\subfigure[Topic fixed effects topic]{\includegraphics[width=\linewidth]{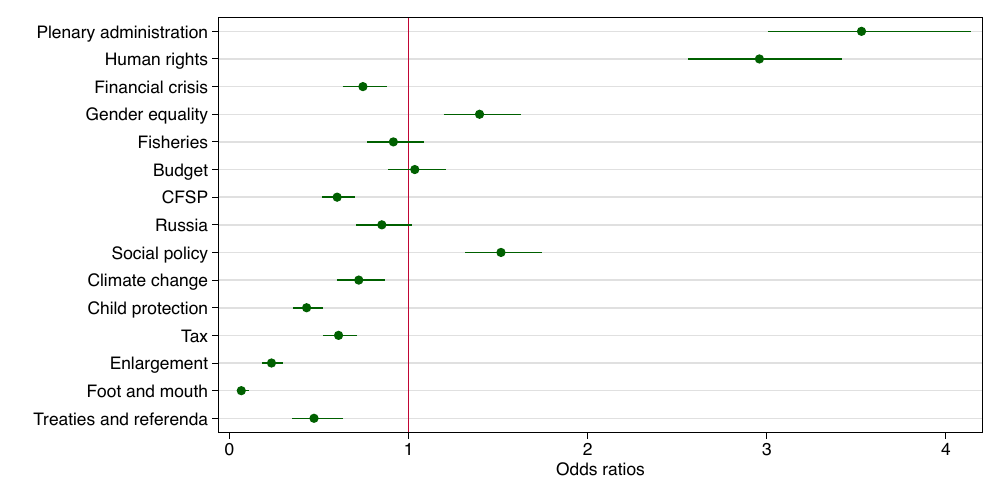}\label{fig:effects1}}
	\vskip 0.4em
	\subfigure[Party group fixed effects]{\includegraphics[width=\linewidth]{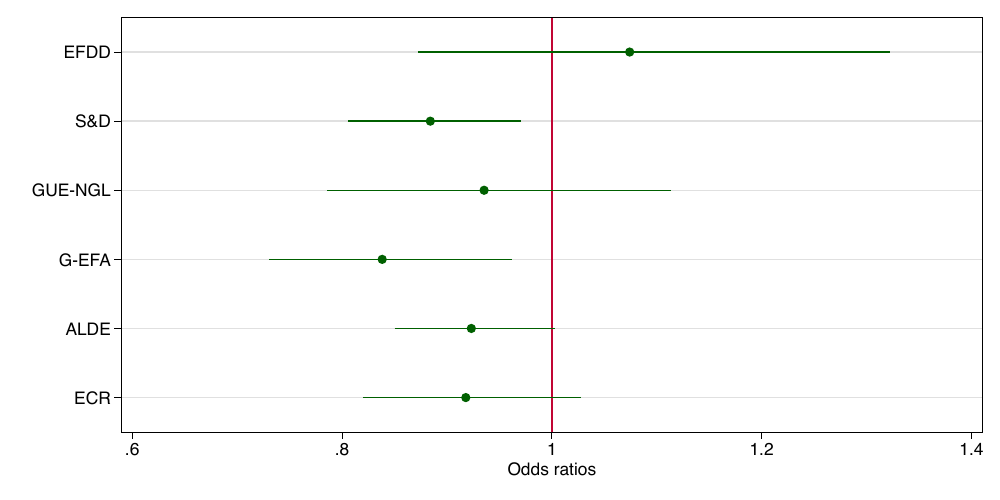}\label{fig:effects2}}
	\vskip -0.5em
	\caption{Fixed effects from regression model.}
	\label{fig:effects}
	\vskip -0.8em
\end{figure}

In order to clarify the substantive size of the effects found in the model, \reffig{fig:effects1} plots the odds ratio of different topics that entered into the regression model but were not displayed in \reffig{fig:regfig}. To plot these fixed effects odds rations, we treat the Euro crisis topic as the baseline. As can be seen, there are significant differences observed between the prevalence of different topics. The most prevalent topic is related to administrative matters in the plenary, and the odds of this topic appearing in an MEP speech are about 3.5 times greater than the odds of a speech relating to the Euro crisis topic. This is not surprising given that administrative matters frame all discussions in the plenary. Perhaps more surprising is the prevalence of the human rights topic relative to the other topics in the analysis. The odds of a speech relating to human rights is about 3 times greater than the odds that a speech relates to the Euro crisis. The relative prevalence of this topic suggests that MEPs regularly comment on human rights issues. Indeed, when one delves into the speeches appearing in this topic, a broad concern for violations of human rights across different contexts is evident. The relative prevalence of topics such as gender equality and social policy is also noteworthy, and suggests that the Parliament actively debates such issues despite the fact that the EU has little formal legislative competencies in these areas.


\reffig{fig:effects2} plots the odds ratios associated with different party groups within the Parliament, treating the European People's Party (EPP) group as the baseline. As can be seen, there is some variation in the odds that a speech on a given topic emerges from a given party, but most of these differences are not statistically significant. This result reflects the fact that speech time is distributed between party groups according to their relative size. Both the \textit{Progressive Alliance of Socialists and Democrats} (S \& D) and the \textit{European Greens--European Free Alliance} (G--EFA) groups are found to differ from the EPP group in terms of the odds a speech on a given topic comes from them. The odds of a topic speech being from either of these groups is less than the odds of a topic speech being from the EPP by a factor of just over 0.1.

\section{Conclusions}
\label{sec:conc}

In this paper, we have proposed a new two-layer matrix factorization methodology for identifying topics in large political speech corpora over time, designed to identify both niche topics related to events at a particular point in time and broad, long-running topics.
We applied this method to a new corpus of all $\approx 210k$ English language plenary speeches from the European Parliament during a 15 year period. In terms of providing substantive insight into the political processes of the European Parliament, the topic modeling method has allowed us to unveil the political agenda in the Parliament, and the manner in which this has evolved over the time period under consideration. By considering three distinct case studies, we have demonstrated the distinctions that can be drawn between the day-to-day political work of the Parliament in policy areas such as fisheries on the one hand, and the manner in which exogenous events such as economic crises and failed treaty referenda can give rise to new topics of discussion between MEPs on the other. Once the Parliamentary agenda was extracted from the corpus of speeches, we explored the determinants of MEP attention to particular topics in the 7th sitting of the Parliament. We demonstrated how MEP ideology and voting behavior affect whether or not they choose to contribute to a topic, and once such a decision has been made, we demonstrated how the committee structure of the Parliament structures MEP contributions on a given topic. 

The initial insights provide by the dynamic topic modeling approach presented in this paper demonstrate the potential of these methods to uncover the latent dynamics in MEP speech-making activities and thus allow for new insights into how the EU functions as a political system. Much remains to be explored in terms of the patterns in political attention that emerge from the topic modeling approach. For instance, one would expect that political attention might well translate into influence over policy outcomes decided upon in the Parliament. Tracing influence to date has been difficult, as a macro-level picture of where and on what topics MEP attention lies has been unavailable. Linking political attention to political outcomes would help to unveil who gets what and when in European politics, which is a central concern for a political system that is often criticized for lacking democratic legitimacy. this is but one direction in which future research might proceed.

While this paper has focused on European Parliament speeches, the proposed topic modeling method has a number of potential applications in the study of politics, including the analysis of speeches in other parliaments, political manifestos, and other more traditional forms of political texts. It is also generally appropriate in domains where large-scale, longitudinal text corpora are naturally represented in discrete segments.



\vskip -0.5em
\noindent {\bf{{Acknowledgments.}}} This research was partly supported by Science Foundation Ireland (SFI) under Grant Number SFI/12/RC/2289.

\bibliographystyle{abbrv}
\bibliography{europarl} 

\balancecolumns
\end{document}